\documentclass[conference]{ieeeconf}
\IEEEoverridecommandlockouts      

\usepackage[table, x11names]{xcolor}
\usepackage[colorlinks=true,linkcolor=black,citecolor=black,urlcolor=blue]{hyperref}
\usepackage{times} 
\usepackage{amsmath} 
\usepackage{amssymb}  
\usepackage{graphicx}
\usepackage{mathtools}
\usepackage{float}
\usepackage{booktabs}
\usepackage[font=footnotesize]{caption}
\usepackage[font=footnotesize]{subcaption}
\usepackage[outdir=./]{epstopdf}
\usepackage{flushend}
\usepackage{footmisc}

\usepackage{balance}

\definecolor{brewerGreen0}{HTML}{E5F5F9}
\definecolor{brewerGreen1}{HTML}{99D8C9}
\definecolor{brewerGreen2}{HTML}{2CA25F}

\definecolor{brewerCyan0}{HTML}{ECE2F0}
\definecolor{brewerCyan1}{HTML}{A6BDDB}
\definecolor{brewerCyan2}{HTML}{1C9099}

\definecolor{brewerGrey0}{HTML}{F0F0F0}
\definecolor{brewerGrey1}{HTML}{BDBDBD}
\definecolor{brewerGrey2}{HTML}{636363}

\definecolor{revisionColor}{HTML}{0238A8} 
\definecolor{lastRevisionColor}{HTML}{CC4C02} 

\usepackage{cite}

\usepackage{balance}
\usepackage{multirow} 

\usepackage{glossaries-extra}
\setabbreviationstyle[acronym]{long-short}
\glssetcategoryattribute{acronym}{nohyperfirst}{true}

\makeglossaries

\newacronym{slam}{SLAM}{Simultaneous Localization and Mapping}
\newacronym{sfm}{SfM}{Structure from Motion}
\newacronym{pgo}{PGO}{Pose-Graph Optimization}
\newacronym{vpr}{VPR}{Visual Place Recognition}
\newacronym{sgd}{SGD}{Stochastic Gradient Descent}
\newacronym{ils}{ILS}{Iterative Least-Squares}
\newacronym{icp}{ICP}{Iterative Corresponding Point}
\newacronym{gn}{GN}{Gauss-Newton}
\newacronym{lm}{LM}{Levenberg-Marquardt}
\newacronym{pcg}{PCG}{Preconditioned Conjugate Gradient}
\newacronym{map}{MAP}{Maximum-A-Posteriori}
\newacronym{gf}{GF}{Gaussian Filters}
\newacronym{pf}{PF}{Particle Filters}
\newacronym{sdp}{SDP}{Semi-Definite Programming}
\newacronym{bst}{BST}{Binary Search Tree}
\newacronym{ndt}{NDT}{Normal Distributed Transform}
\newacronym{ba}{BA}{Bundle Adjustment}

\def\secref#1{Sec.~\ref{#1}}
\def\figref#1{Fig.~\ref{#1}}
\def\tabref#1{Tab.~\ref{#1}}
\def\eqref#1{Eq.~(\ref{#1})}

\def\lidar{LiDAR}
\def\lidars{LiDARs}
\def\rgbd{RGB-D}

\usepackage{amsopn}

\newcommand{\bv}{\mathbf{v}}

\newcommand{\bt}{\mathbf{t}}

\newcommand{\bK}{\mathbf{K}}

\newcommand{\bI}{\mathbf{I}}
\newcommand{\bX}{\mathbf{X}}

\newcommand{\bR}{\mathbf{R}}

\newcommand{\bJ}{\mathbf{J}}

\newcommand{\map}{\zeta}

\newcommand{\depth}{d}
\newcommand{\XSet}{\bX_{1:N}}
\newcommand{\optSet}{\XSet^*}

\newcommand{\bbR}{\mathbb{R}}
\newcommand{\bbSE}{\mathbb{SE}}
\newcommand{\bbSO}{\mathbb{SO}}

\newcommand{\lieSE}{\mathfrak{se}}

\newcommand{\be}{\mathbf{e}}

\newcommand{\bq}{\mathbf{q}}

\newcommand{\bx}{\mathbf{x}}

\newcommand{\bu}{\mathbf{u}}
\newcommand{\bn}{\mathbf{n}}

\newcommand{\bp}{\mathbf{p}}

\newcommand{\bDelta}{\mathbf{\Delta}}

\newcommand{\bDeltax}{\mathbf{\Delta x}}

\newcommand{\bDeltaX}{\mathbf{\Delta X}}
\newcommand{\bDeltat}{\mathbf{\Delta t}}
\newcommand{\bDeltaR}{\mathbf{\Delta R}}
\newcommand{\bzero}{\mathbf{0}}

\newcommand{\bOmega}{\mathbf{\Omega}}

\DeclareMathOperator*{\argmin}{argmin}
\DeclareMathOperator*{\atantwo}{atan2}

\def\g2o{$g^2o$}
\def\t2v{\mathrm{log}}
\def\v2t{\mathrm{exp}}
\def\ev2t{\mathrm{ev2t}}

\def\skew#1{{\lfloor{#1}\rfloor}_\times}

\newcommand{\I}{\ensuremath{\mathcal{I}}}
\newcommand{\predI}{\ensuremath{\mathcal{\hat I}}}
\newcommand{\channel}{\mathrm{c}}

\newcounter{todonum}

\usepackage{lipsum}

\title{\LARGE \bf Photometric LiDAR and RGB-D Bundle Adjustment}

\author{Luca Di Giammarino \and Emanuele Giacomini \and Leonardo Brizi \and Omar Salem \and Giorgio Grisetti

\thanks{All authors are with the Department of Computer, Control, and Management Engineering "Antonio Ruberti", Sapienza University of Rome, Italy,
Email:\,\,{\tt\footnotesize{\{digiammarino, giacomini, brizi,
salem, grisetti\}@diag.uniroma1.it.}}}%

}%

\begin{document}
\maketitle		

\begin{abstract}
The joint optimization of the sensor trajectory and 3D map is a crucial characteristic of \gls{slam} systems. To achieve this, the gold standard is \gls{ba}. Modern 3D \lidars~now retain higher resolutions that enable the creation of point cloud images resembling those taken by conventional cameras. Nevertheless, the typical effective global refinement techniques employed for \rgbd~sensors are not widely applied to \lidars. This paper presents a novel \gls{ba} photometric strategy that accounts for both \rgbd~and \lidar~in the same way. Our work can be used on top of any SLAM/GNSS estimate to improve and refine the initial trajectory. We conducted different experiments using these two depth sensors on public benchmarks. Our results show that our system performs on par or better compared to other state-of-the-art ad-hoc SLAM/\gls{ba} strategies, free from data association and without making assumptions about the environment. In addition, we present the benefit of jointly using \rgbd~and \lidar~within our unified method. We finally release an open-source CUDA/C++ implementation \footref{repo}.    
\end{abstract}
	
\section{Introduction}
\label{sec:intro}
\gls{slam} has gained significant popularity in the field of robotics,
and thanks to approximately three decades of research, there are now
effective solutions available. As \gls{slam} is widely used in various
sectors, such as autonomous driving, augmented reality, and space
exploration, it continues to attract significant attention from
academia and industry. Modern SLAM systems typically
comprise two key components: a front-end that estimates sensor
odometry in real-time and a back-end that optimizes previous sensor
poses and, eventually, the 3D map. The gold standard for the back-end
optimization is \gls{ba} \cite{triggs2000bundle}. Photometric (or
direct) approaches have been used by the computer vision community to
tackle the \gls{slam} or \gls{sfm} problems. The direct techniques
address registration by minimizing the pixel-wise error
between image pairs. By not relying on specific features and having
the potential of operating at subpixel resolution on the entire image,
direct approaches do not require explicit data association and boost
the registration accuracy \cite{kerl2013dense, schops2019bad}.
Although these methods have been successfully employed on monocular,
stereo, or \rgbd~sensors, their use on 3D \lidar~data has not been
widely exploited. Della Corte \emph{et al.}
\cite{della2018general} presented a general photometric registration
methodology that extends direct approaches to different projective
models and enhances the optimization robustness by considering
additional channels such as normals. Nowadays, 3D \lidars~offer up to
128 channels of resolution, measuring more than 5M points per
second. The enhancement of density, the increment in vertical
resolution, and the capacity to measure the reflectivity of a surface
make it logical to render a scan onto a panoramic image. This
consideration makes photometric approaches attractive for 3D \lidars.
 
\begin{figure}[t]
	\centering
        \includegraphics[width=1\linewidth]{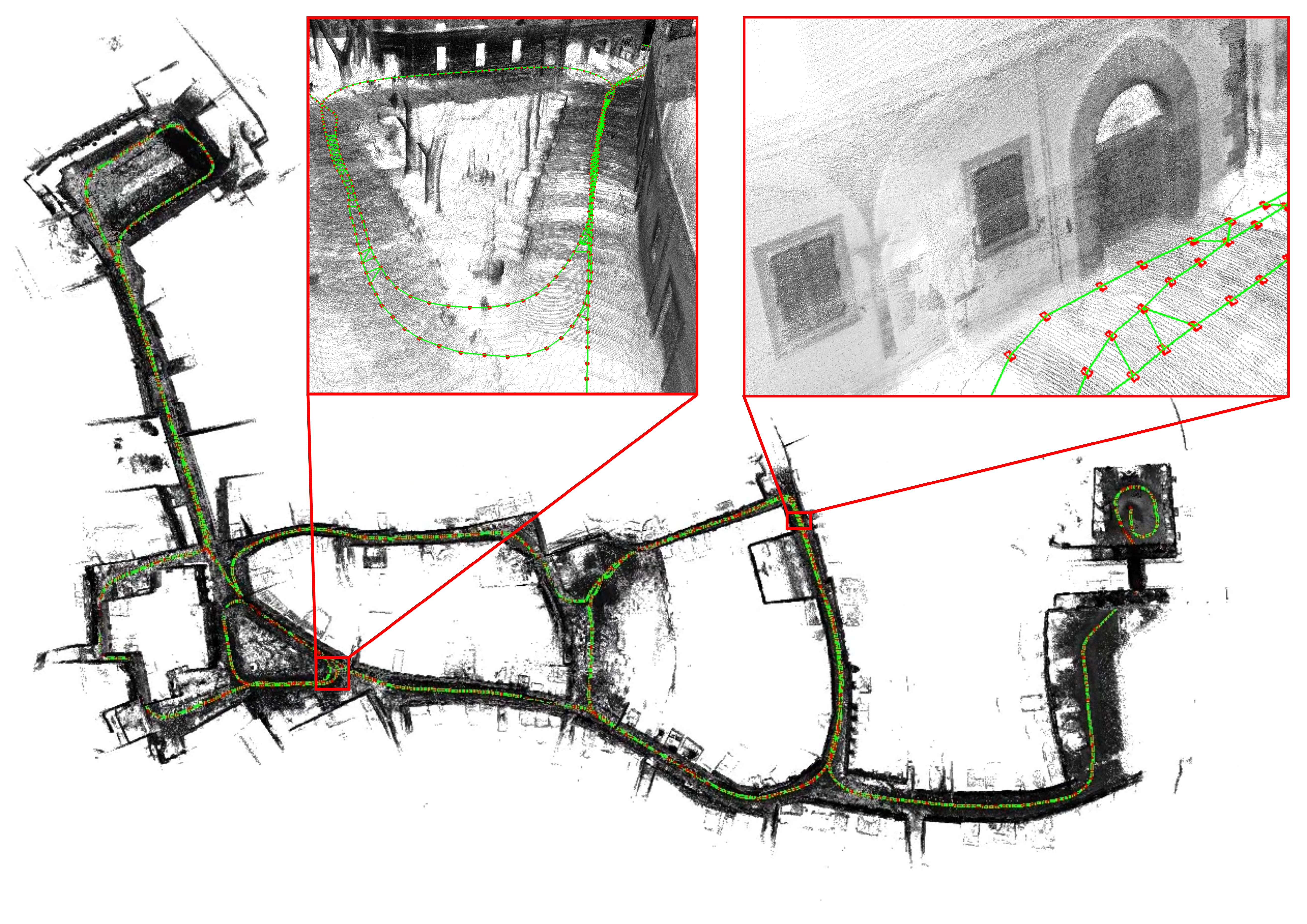}
	\caption{Reconstruction of Viterbo city-center (Italy) using
		our data recorded with an OS0-128. The trajectory, which is
		about 2 km long, has been estimated first with MD-SLAM
		\cite{di2022md} and then refined with our photometric \gls{ba}
		strategy. This image highlights both the global and local
		consistencies. We show reconstruction details
		with multiple scans of the same places acquired over time.}
	\label{fig:motiv}
\end{figure}

The main contribution of this paper is a unified photometric \gls{ba}
strategy that works for both \rgbd~and \lidar. Our method aims to refine the trajectory coming from a SLAM/GNSS system to
maximize its photometric consistency. Our approach implicitly
addresses the data association and straightforwardly supports multiple
heterogeneous sensors. We performed a comparative evaluation of
benchmark data concerning state-of-the-art sensor-specific
refinement strategies and SLAM algorithms. Results show that our
simple optimization schema is very effective, performing on par or
better than methods specialized for RGB-D and LiDAR data. We also demonstrate how our photometric \gls{ba} strategy can be
improved by fusing 3D \lidar~and \rgbd. We release an open-source
CUDA/C++ implementation of this
work \footnote{\url{https://github.com/digiamm/ba\_md\_slam} \label{repo}}.

\figref{fig:motiv} shows a reconstruction of the Viterbo city-center
(Italy) using our self-recorded data. From the detailed views, it is
possible to appreciate the fine map resolution after performing our
\gls{ba} strategy.

\section{Related Work}
\label{sec:related}
Approaches for global refinement such as \gls{ba} are
widely used in Visual SLAM and \gls{sfm} systems but are less
common for LiDAR. In this work, we provide a unified photometric
global registration method for both \rgbd~and \lidar~that can improve the accuracy of the
trajectory - and hence the map - obtained by standard SLAM systems that
rely on \gls{pgo}. \gls{pgo} formulation reduces the optimization
problem's size but approximates the original problem by
marginalizing the projective observations.

In the \rgbd~SLAM field, \gls{ba} can be classified into two
categories: \textit{direct} and \textit{indirect}. In the following,
we will explain the difference between these two paradigms and review
the state-of-the-art of both approaches. We finally discuss 
\gls{ba} applications in \lidar~SLAM.

Indirect methods are based on feature detection and matching between
images. The camera poses and the 3D structure are then estimated by
minimizing the reprojection error of this set of feature points. These
methods are preferred in SLAM applications because they are
faster since they operate on much fewer data. The data reduction,
employed by extracting features, renders indirect methods less
sensitive to calibration and synchronization issues
\cite{schops2019bad}. State-of-the-art indirect SLAM implementations perform windowed \gls{ba} to refine the map under construction in a neighborhood of the current sensor location. Global \gls{ba} is invoked upon loop closures on
the entire trajectory. To keep the size of the problem tractable during global \gls{ba}, the trajectory is subsampled in a set of keyframes, and only some salient feature points are considered in the optimization~\cite{mur2017orb, klein2007parallel}.  

Direct methods, on the other hand, also known as photometric, do not rely on feature detection and matching. Instead, they directly
minimize the photometric error between overlapping images using all
information available in the image. More specifically, this error is
calculated as the difference between the measured and the predicted pixel intensities. Computing such a prediction relies on an estimate
of the 3D structure captured by the camera and its parameters.
Delaunoy and Pollefeys propose an offline dense 3D reconstruction
methodology, in which refinement of the 3D scene and camera
parameters is performed simultaneously, with the primary goal of
minimizing photometric error \cite{delaunoy2014photometric}. The scene
is represented by triangular meshes, requiring frequent remeshing
during the optimization. This makes the approach computationally
demanding. Goldl{\"u}che \emph{et al.} presents a new variational framework
that enables the precise representation of a 3D scene by estimating a
super-resolution curved surface. Unlike the previous work using
typical triangular meshes, the authors utilize a smooth surface
\cite{goldlucke2014super}. Decoupling the camera motion from the
optimization process leads to a convex objective functional, resulting
in an improved estimate and high-quality texture output. Slavcheva
\emph{et al.}, instead of building a mesh, perform pairwise alignment
registering directly two consecutive signed distance functions (SDFs)
generated from the depth images. This kind of registration is then
used also as global refinement \cite{slavcheva2016sdf}. The direct
methods presented previously are computationally heavy, being mainly
used for offline 3D reconstructions but not suitable for on-line
estimation.

To enhance the computation, some researchers started to exploit the
decomposability of the problem by leveraging its structure. Others
investigated data reduction strategies to reduce the problem's size while preserving
enough information to obtain an accurate estimate.
To reduce the
problem size, Hatem Alismail \emph{et al.} leverage the fact that most
of the image points do not contribute equally to the optimization
and consider in the error function only pixels with reasonable
gradient \cite{alismail2017photometric}.

To reduce computational demand for large scale \gls{ba}
problems, Eriksson \emph{et al.}  propose a consensus-based
optimization to parallelize Bundle Adjustment in \gls{sfm}
applications \cite{eriksson2016consensus}. In a more recent
development, Demmel \emph{et al.} proposed a novel solution in
Distributed \gls{ba} \cite{demmel2020distributed}.  Specifically, the author
broke down the original problem into smaller, more manageable subparts
using the k-means clustering method. To allow more efficient
processing, the resulting subgraphs are structured into relative
well-constrained connected segments.

In between the class of direct and indirect methods, we find the work of
Forster~\emph{et al.}~\cite{forster2016svo}, which proposes a hybrid
method to estimate the camera's motion. First, an initial
guess of the sensor location is computed by minimizing the reprojection error
of the world points. Then the estimate is refined by minimizing
the photometric error of the patches around the feature points. The
final map refinement step is done by performing direct \gls{ba}.

In the last years, thanks to the technology enhancement, the community
focused on embedding \gls{ba} refinement in SLAM applications. To combine the accuracy of direct methods with the robustness of feature-based ones, hybrid approaches gained traction. These methods mix both direct and indirect error terms in their optimization strategies. One
such instance is Bundle Fusion \cite{dai2017bundlefusion}, which
refine the global estimate by interleaving feature-based and
photometric \gls{ba}. The photometric refinement does not take into
account the structure, but only the camera poses. Similarly, in
BAD-SLAM the motion estimation and global refinement processes are
accomplished by minimizing a cost function that accounts for geometric
and photometric errors \cite{schops2019bad}. The global refinement
process is broken down into three main steps: first, the 3D scene
modeled by surfel is refined, then the camera poses are optimized,
keeping the model fixed, and finally, the camera's intrinsics are
refined.

In parallel, the community approached LiDAR-based SLAM by seeking
alternative representations for the dense 3D point clouds. Given the
accuracy of these measurements, the robotics community addressed
the problem of building a map incrementally registering new scans.

Many registration techniques have been exploited using
\lidar~data. These include 3D salient features
\cite{zhang2014loam,serafin2016fast}, subsampled clouds
\cite{velas2016collar} or \gls{ndt} \cite{stoyanov2012fast}. Nowadays,
LiDAR Odometry and Mapping (LOAM) is perhaps one of the most popular
methods for LiDAR odometry \cite{zhang2014loam, zhang2015visual}. It
extracts distinct features corresponding to surfaces and corners, then
used to determine point-to-plane and point-to-line distances to a
voxel grid-based map representation. A revised approach (Lego-LOAM)
has been suggested \cite{shan2018lego}, which takes advantage of a
ground surface in its segmentation and optimization steps. Odometry
estimation techniques, or more generally 3D point clouds registration
routines, coupled with place recognition and \gls{pgo} show satisfying
results within \lidar~SLAM \cite{shan2018lego, di2022md}. This makes
\gls{pgo} the gold standard optimization method in \lidar~community.
A pose-graph represents the trajectory, and observations between pairs
of poses along the path are computed by registering the two
overlapping clouds. Hence, an observation is a relative transform and
potentially a covariance matrix. With this approximation, the
constraints between the poses can be represented in a relatively
compact manner. The optimum of a \gls{pgo} is the configuration of
poses that is maximally consistent with the transforms in the
measurements. Albeit efficient, \gls{pgo} approaches operate on approximating the original problem since pairwise measurements are
computed once during the SLAM phase and never revised. Unavoidable
drifts will accumulate, and wrong behavior of the place recognition
might lead to inconsistent graphs as shown
in~\cite{zhang2016degeneracy}.

In order to remove these inconsistencies, recently Liu and Zhang
presented a global geometrical optimization methodology that considers
cloud measurements. In particular, it formulates a cost function
based on LOAM features (i.e., edges and planes) and globally optimizes
the trajectory to maximize the features' overlap. This approach
performs a static data association based on the co-visibility of
landmarks; thus, it requires a good initial guess to operate.

In
recent years, the vertical resolution of modern 3D \lidars~ increased,
and these devices provide scans that resemble more and more dense
panoramic images. This allows us to transfer results about direct
global refinement from vision to \lidar. In this work, we present a
unified \gls{ba} strategy that works independently for \lidar~and
\rgbd~in the same way. By operating directly on images, our
method constantly refines the data association during the
optimization; hence it is less sensitive to poor initial guesses. Our
algorithm's capabilities are demonstrated through quantitative
and qualitative analyses, consistently improving the initial estimates
provided by any SLAM systems.

\section{Basics}
\label{sec:basics}
Our method models the problem of photometric \gls{ba} as an optimization problem.
In this section, we present some notation and review some
concepts used in the remainder of the paper: parameterization
of the transformations and projective models for \rgbd~and \lidar{}.

We parameterize the sensor motion to the class of rigid body motions
forming the special euclidean group $\bbSE(3)$. A common
representation for rigid body motions is a transformation matrix $\bX$,

\begin{equation}
	\bX = \begin{bmatrix}
	\bR & \bt\\
	\bzero & 1 \\
	\end{bmatrix}, \qquad 
	\bX^{-1} = \begin{bmatrix}
	\bR^T & -\bR^T\bt\\
	\bzero & 1 \\
	\end{bmatrix},
\end{equation}
here $\bR \in \bbSO(3)$ is a $3 \times 3$ rotation matrix and $\bt$ is a 3D
translation vector.

Our method deals with both \lidar~and \rgbd~cameras homogeneously. The only adaptation to be done for a specific sensor is to
define the appropriate projective model. These two models are
illustrated in the remainder of this section. 

A projection is a
mapping $\pi : \bbR^3 \rightarrow \Gamma \subset \bbR^2$ from a world
point $\bp = [ x, y, z ]^T$ to image coordinates
$\bu = [u, v]^T$.  Knowing the depth
or the range $\depth$ of an image point $\bu$, we can calculate the
inverse mapping $\pi^{-1} : \Gamma \times \bbR \rightarrow \bbR^3$,
more explicitly $\bp = \pi^{-1}(\bu, \depth)$. We will refer to this
operation as unprojection. We remind that in the case of \lidar, $d$
represents the \textit{range}, namely the distance between the endpoint and the origin of the observer. Differently for \rgbd, $d$ represents the \textit{depth}, which is the distance between the endpoint and the image plane. For compactness, we will only refer to $d$ as depth in the remainder of the paper.

\textbf{Pinhole Model (\rgbd):} Let $\bK$ be the camera matrix. Then, the
pinhole projection of a point $\bp$ is computed as
\begin{align}
	\pi_p(\bp) &= \phi(\bK \, \bp), \label{eq:pinhole-projection}\\
	\bK&= \begin{bmatrix}
	f_x& 0 & c_x\\
	0 & f_y & c_y\\
	0 & 0 &1
	\end{bmatrix}, \label{eq:camera-matrix-p}\\
	\phi(\bv) &= \frac{1}{v_z}
	\begin{bmatrix}
	v_x \\ v_y
	\end{bmatrix},
\end{align}
with the intrinsic camera parameters for the focal length~$f_x$, $f_y$
and the principle point~$c_x$, $c_y$. The function~$\phi(\bv)$ is the
homogeneous normalization with $\bv=[v_x,v_y,v_z]^T$.
	
\textbf {Spherical Model (\lidar):} Let $\bK$ be a camera matrix in the form
of~\eqref{eq:camera-matrix-p}, where $f_x$ and $f_y$ specify
respectively the resolution of azimuth and elevation and $c_x$ and
$c_y$ their offset in pixels. The function $\psi$ maps a 3D point to
azimuth and elevation. Thus the spherical projection of a point is
given by
\begin{align}
	\pi_s(\bp) &= \bK_{[1,2]} \psi(\bp), \label{eq:spherical-projection}\\
	\psi(\bv) &= 
	\begin{bmatrix}
	\atantwo(v_y, v_x) \\
	\atantwo\left(v_z, \sqrt{v_x^2 + v_y^2}\right)\\ 1
	\end{bmatrix}.
\end{align}
In the spherical model $\bK_{[1,2]} \in \bbR^{2 \times 3}$, being the
third row in $\bK$ suppressed.
        
\section{Our Approach}
\label{sec:main}
\begin{figure}[h]
	\centering
	\includegraphics[width=0.99\linewidth]{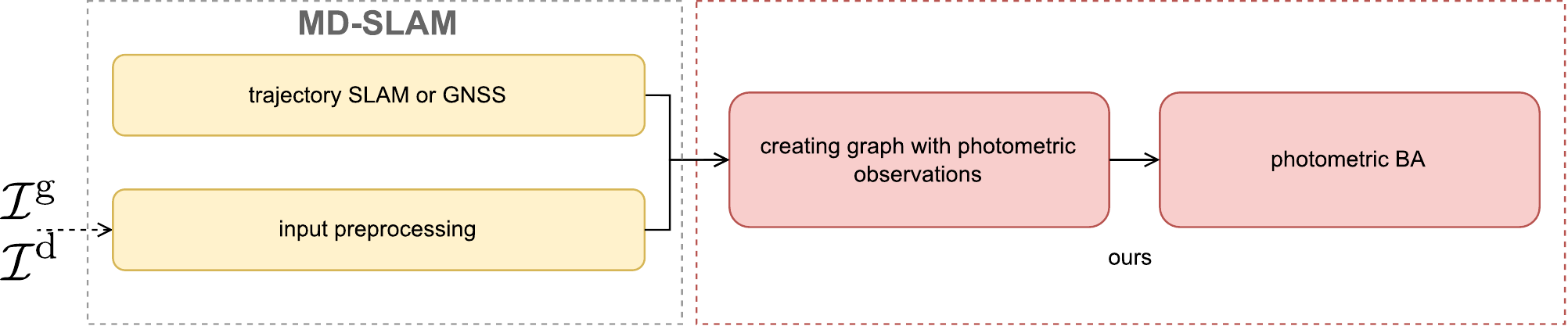}
	\caption{The flow of our approach. The input of our system contains the initial guess of the sensor pose, the intensity image $\I^\mathrm{g}$
		and the depth image $\I^\mathrm{d}$. MD-SLAM already provides the input in the correct format since the preprocessing step is embedded in the SLAM system. The core of our refinement strategy consists in creating a graph with photometric observations (\secref{sec:graph-aug}) and running a global optimization on the set of poses $\bX_{1:N}$, as detailed in \secref{sec:error-min}.}
	\label{fig:flow}
\end{figure}

The goal of our approach is to compute the set of sensor poses $\bX_{1:N}$
whose photometric error between overlapping frames is minimized. The
input of our system is a set of triplets
$\left<\bX_i,\I^\mathrm{g},\I^\mathrm{d}\right>$, containing the
initial guess of the sensor pose, the grayscale/intensity image $\I^\mathrm{g}$
and the depth image $\I^\mathrm{d}$.  The workflow of our method is illustrated in \figref{fig:flow}.  The first step is to generate an augmented image
pyramid from each pair of images in a triplet (\secref{sec:pyramids}).
The second step is determining which pairs of images observe a common
structure, which is discussed in \secref{sec:graph-aug}.  Finally, these
pairs will be used to instantiate a photometric optimization problem
presented in \secref{sec:error-min}.  Without using geometrical
error functions or structure optimization, our approach is entirely
free from any feature association. This makes our method simple,
compatible with multiple depth sensors, and competitive with ad-hoc
state-of-the-art approaches.

\subsection{Input Preprocessing}
\label{sec:pyramids} 
The input of our photometric refinement method is a pair of intensity
$\I^{\mathrm{g}}$ and depth $\I^{\mathrm{d}}$ images for each sensor pose.  The output of the preprocessing step is a five-channel image pyramid for each input pair. The first two channels of an image in the pyramid are intensity and depth, while the other three channels encode the surface normals. To calculate the normal at pixel
$\bu$ we unproject the pixels in the neighborhood $\mathcal U=\{\bu':
\|\bu-\bu'\| < \tau_\bu\}$ of a radius $\tau_\bu$ inversely
proportional to the range at the pixel $\I^{\depth}(\bu)$.
The normal $\bn_\bu$ is the one of
the plane that best fits the unprojected points from the set $\mathcal
U$, and is oriented towards the observer.
All valid normals are assembled in a normal image
$\I^\mathrm{n}$, so that $\I^\mathrm{n}(\bu)=\bn_\bu$.
Hence, the final five-channel image is obtained by stacking together $\I^\mathrm{g}$, $\I^\mathrm{d}$, and $\I^\mathrm{n}$.
In the remainder, we will refer to the generic channel as a \emph{cue} $\I^\mathrm{c}$.

Photometric approaches perform an implicit data association at a pixel
level. Whereas attractive for their accuracy, these methods suffer from
relatively small convergence basins that decrease with the image's resolution: the higher the image resolution, the narrower the convergence basin
will be. To lessen this effect, we generate multiple copies of the same image at decreasing resolution to form a pyramid.
The optimization will proceed from the coarser to the finest level (\figref{fig:hopt}).

Each level of a pyramid consists of a multi-cue image generated from $\I^\mathrm{g}$, $\I^\mathrm{d}$ and $\I^\mathrm{n}$, by downscaling at user-selected resolutions. In our experiments, we typically use three scaling levels, each half the resolution of the previous level.

\subsection{Graph Construction}
\label{sec:graph-aug}
The input of our system is a set of triplets
$\left<\bX_i,\I^\mathrm{g},\I^\mathrm{d}\right>$, containing the
initial guess of the sensor pose and the intensity/depth image pairs.
To instantiate \eqref{eq:total-error-multicue} we need to determine the matching pairs
$\left<i,j\right>$. The problem can be visualized as an undirected graph, where each node is a triplet, and an edge between two nodes encodes a potential match. 

\begin{figure}[H]
	\centering
	\includegraphics[width=0.99\linewidth]{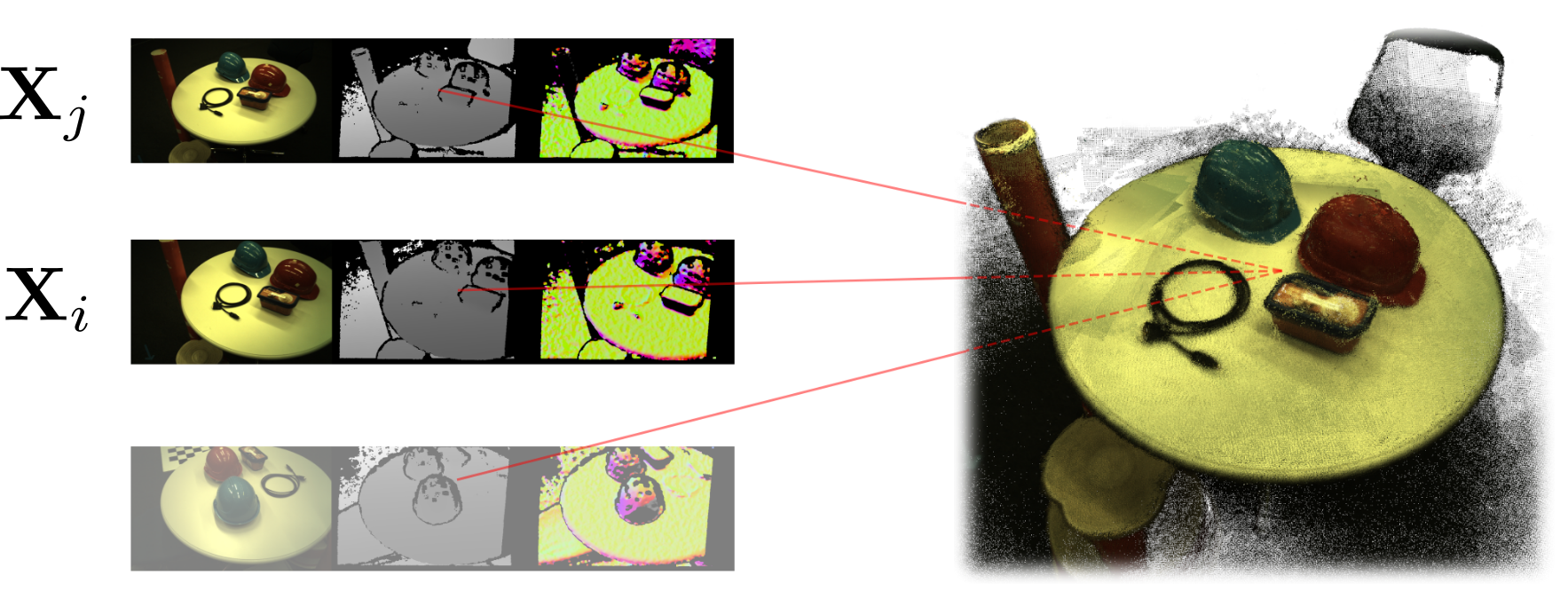}
	\caption{An example of pose-pairs associated, our \gls{ba} strategy relies on the association of $\bX_i$ and $\bX_j$ if they share observations. In the picture above, the input images with depth and normals are on the left, and on the right is the reconstructed model using our methodology. Note that in reality, inputs of our \gls{ba} are pyramids (as discussed in \secref{sec:pyramids}, i.e., images of different resolutions). Here, we show just one level for simplicity. The data used is from ETH3D.}
	\label{fig:augm-graph}
\end{figure}

To compute the matches, we start from the initial assignment of poses $\{\bX_n\}$  and add edges
to the graph based on the input data. To this extent, we use a straightforward criterion that generates a matching pair if two poses $\bX_i$, $\bX_j$ are close in space, and their orientations are similar. More specifically,
we create a pair between $\bX_i$, $\bX_j$ if all these conditions
are satisfied:
\begin{enumerate}
	\item the angle between the poses is below a threshold (typically $30 \deg$);
	\item the translation between the poses is below a threshold (typically below $1$ meter);
	\item the ratio of reprojected valid points from $\bX_i$ onto $\bX_j$ is sufficiently high (typically $1/3$).
\end{enumerate}
In addition to these criteria, if the data come from a sequential acquisition, we add matches between subsequent triplets to
model odometry-like constraints. An example of pose-pair associated is shown in \figref{fig:augm-graph}.

\subsection{Photometric Error Minimization}
\label{sec:error-min}
Our method seeks to find the set of transformations $\optSet \in
\bbSE(3)^N$ that minimizes the photometric error between each
candidate pair of sensor poses that observe a common portion of the
environment.  Let $\mathcal I_i$ and $\mathcal I_j$ be two images
acquired from poses $\bX_i$ and $\bX_j$ that form a
matching pair.  Let $\mathcal{I}(\bu)$ be the value of the pixel
$\bu$ in the image $\mathcal I$.

The photometric error at image coordinates $\bu$ in the matching pair is the difference between
$\mathcal I^\mathrm{g}_i(\bu)$ and the pixel $\mathcal I^\mathrm{g}_j(\bu')$ of the second image.
The evaluation point $\bu'$ is computed by unprojecting the pixel $\bu$ from $\I_i^\mathrm{g}$ onto the image plane
of $\I_j^\mathrm{g}$. This accounts for the relative transform $\bX_{j,i} =  \bX_j^{-1} \bX_i$  between the two frames, as follows:
\begin{equation}
\bu' = \pi \left ( \bR_j^T \left ( \bR_i \pi^{-1} \left(\bu,\depth \right) + \bt_i   - \bt_j \right ) \right ).  
\label{eq:uprime}
\end{equation} 
To carry out this operation, the depth at the pixel $\depth = \I^{\depth}(\bu)$ needs to be known.
Standard photometric optimization seeks to find the following minimum:
\begin{equation}
        \optSet=
	\argmin_{\XSet} \sum_{\left<i, j\right>} \sum_{\bu}
	\| 
	\I_i^\mathrm{g}(\bu) - \I_j^\mathrm{g}(\bu')
        \|^2.
        \label{eq:total-error}
\end{equation}
Here the inner summation computes the photometric error of a matching pair ${\left<i, j\right>}$ as the squared norm of the error of all pixels $\bu$
while the outer summation spans over all matching image pairs.

\begin{figure}[h]
	\centering
	\includegraphics[width=0.99\linewidth]{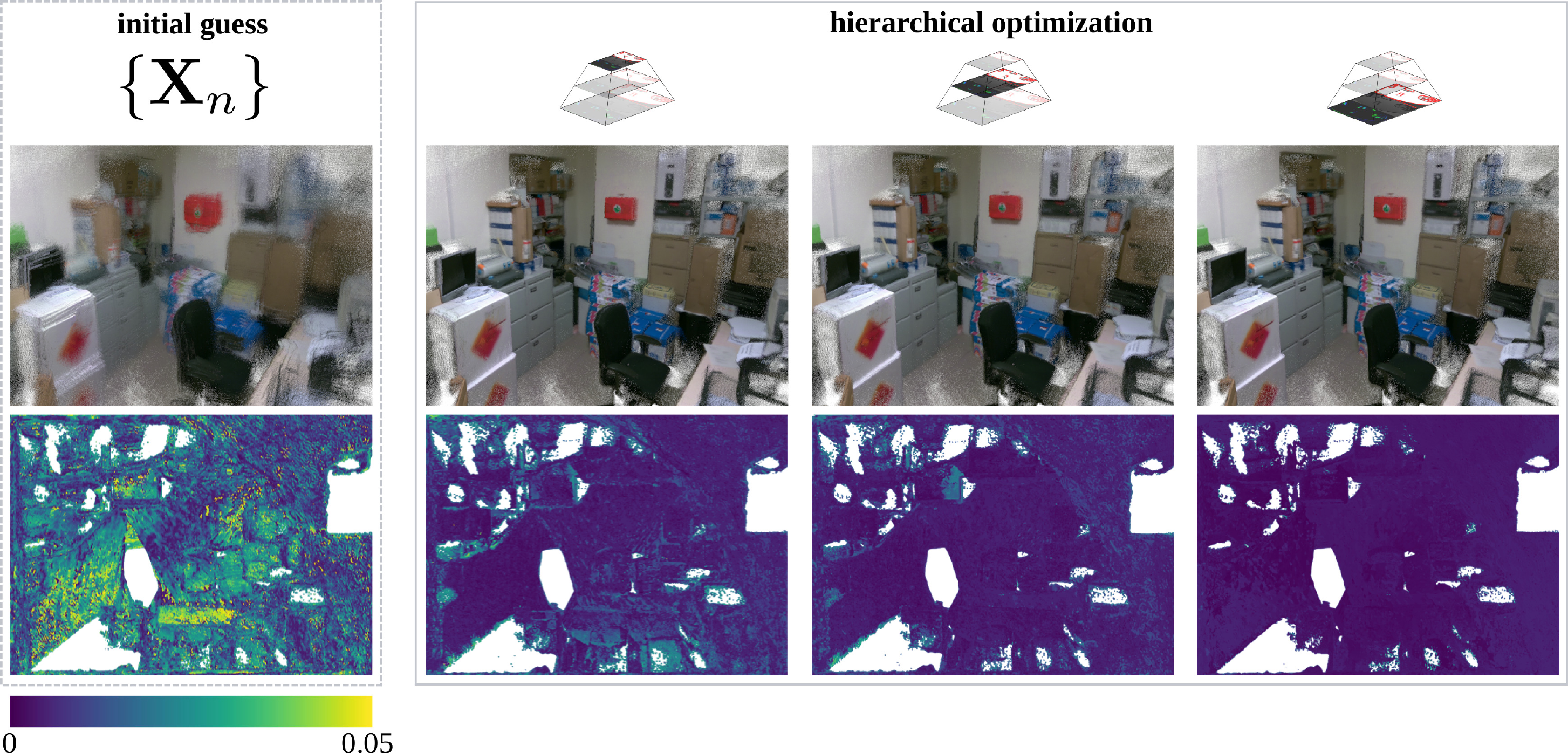}
	\caption{The effect of hierarchical optimization when performing BA. From left to right, we show how the quality of the estimate improves, starting from a bad initial guess. The heatmap is normalized between 0 and 0.005 [m] for better visualization. The data is self-recorded using an Intel D455.}
	\label{fig:hopt}
\end{figure}

\eqref{eq:total-error} models classical photometric error
minimization assuming that the cues are unaffected by $\bX_{j,i}$.
Whereas this is true when operating with pure intensity/grayscale values, normals, and depths change when mapped from the frame $\bX_i$ to the frame $\bX_j$.
As in in~\cite{della2018general} these mappings can be incapsulated by the function $\map^\channel(\bX_{j,i}, \I_i^\channel(\bu))$ that calculates the \emph{pixel}
value of the $\channel^\mathrm{th}$ cue after applying the transform
$\bX_{j,i}$ to the original channel value $\I_i^\channel(\bu)$.
By extension, let $\predI_i ^\channel = \map^\channel(\bX_{j,i}, \I_i^\channel)$ be the image obtained by remapping  $\mathcal I^\channel_i$, according to $\bX_{j,i}$.
We can thus rewrite a more general form of \eqref{eq:total-error} that
accounts for all cues and captures this effect as follows:
\begin{align}
  F(\XSet)&= \sum_{i, j}  \sum_\bu \rho \left \lVert \sum_ \channel
  \predI_i^\channel(\bu) - \I_j^\channel(\bu') \right \rVert^2_{\bOmega^\channel}
  \label{eq:total-error-multicue}
  \\
  \optSet&= \argmin_{\XSet} F(\XSet)
\label{eq:error-min-multicue}
\end{align}
where $\rho$ is a Huber robust M-estimator. More details about \eqref{eq:total-error-multicue} can be found in our supplementary material \footref{repo}.  To carry on the
minimization in \eqref{eq:error-min-multicue} we employ the
Levenberg-Marquardt algorithm implemented in the
\texttt{srrg2\_solver} \cite{grisetti2020least}. At each iteration,
we suppress the occluded portions of the images before evaluating \eqref{eq:total-error-multicue}. The optimization proceeds by seeking
the optimum of all poses, starting from the coarser level. Once
convergence is reached at one level, our system switches to the next
finer one, and the optimization proceeds by choosing the
solution computed so far as an initial guess.  \figref{fig:hopt} shows the effect of this
hierarchical approach applied to a \gls{ba} problem. Generally, the worse the initial guess, the more the required levels.

%

%
%
	

\begin{figure*}[h]
	\centering
	\includegraphics[width=0.99\textwidth]{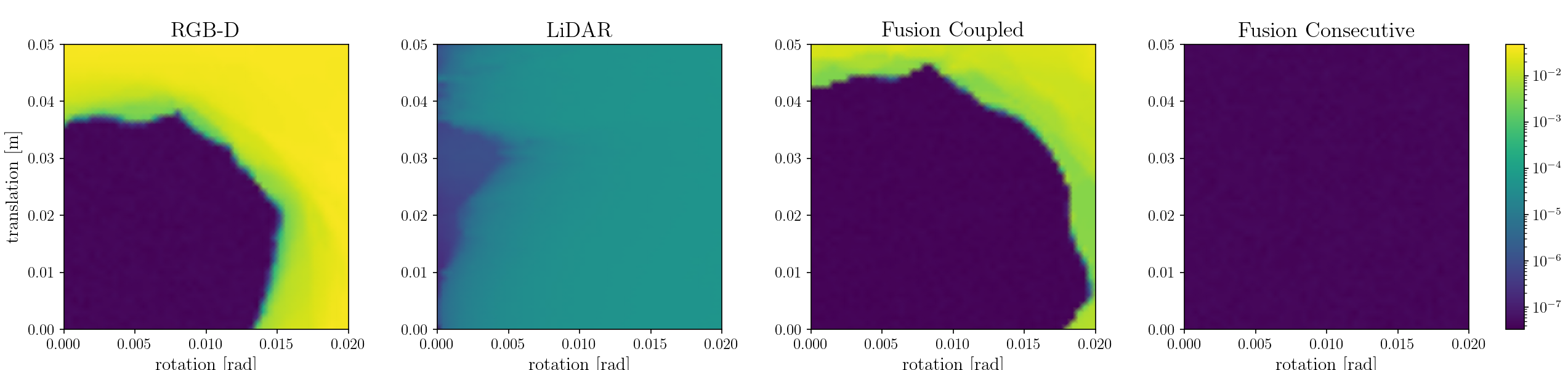}
	\caption{Results of fusion experiments. Plots show how the initial perturbation affects the convergence basin and the algorithm's accuracy. On the x-y axis, respectively, translation [m] and rotation [rad] perturbations, the value is denoted by the mean between rotation and translation error in $\log$ scale. Fusing the two sensors with our straightforward approach always shows better results compared to single sensors.}
	\label{fig:fusion}
\end{figure*}

\section{Experimental Evaluation}
\label{sec:exp}
In this section, we report the results of our method on different
public benchmark datasets. To the best of our knowledge, our approach
is the only open-source photometric \gls{ba} strategy that can deal
with \rgbd~and~\lidar~in a unified manner. Therefore, to evaluate our
system, we compare it with state-of-the-art SLAM and \gls{ba} packages
developed specifically for each of these sensor types.
 
\begin{figure}[t]
	\centering
	\includegraphics[width=0.99\linewidth]{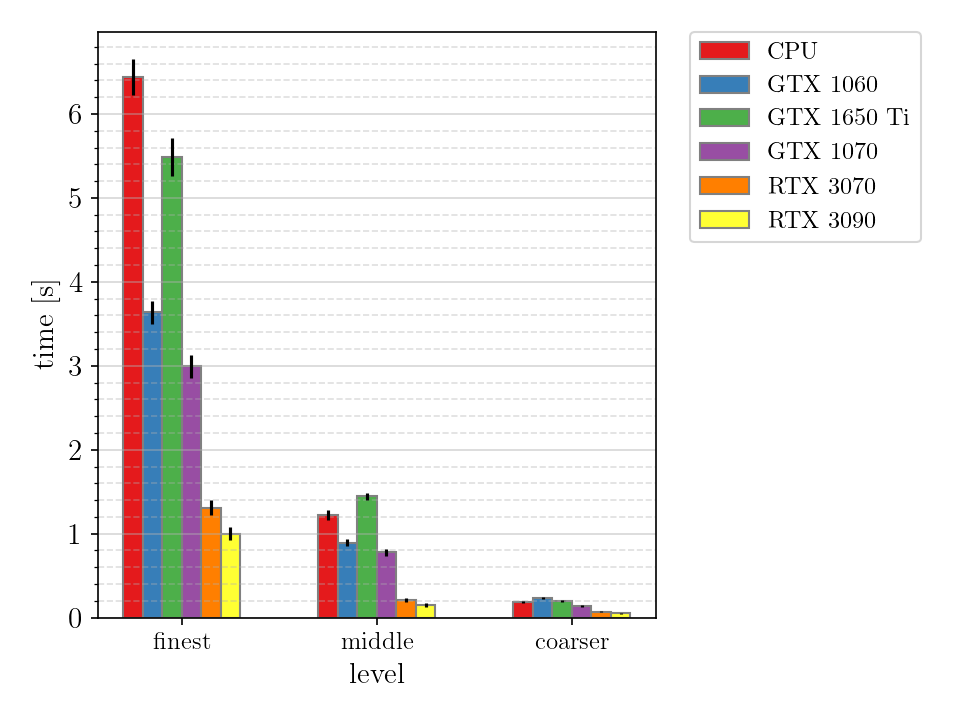}
	\caption{Runtimes of our \gls{ba} strategy evaluated on multiple commercial GPUs and our Intel Core i7-7700K CPU. Specifically, the value is represented by the cumulative time reached to complete an iteration during the optimization process, while the error bar indicates the standard deviation. In this experiment, the finest level includes around 86M pixels, the middle level includes 22M pixels, and the coarser level has around 5M pixels. Time is expressed in seconds.}
	\label{fig:gpus}
\end{figure}

To run the experiments, we used a PC with an Intel Core i7-7700K CPU @
4.20GHz, 32GB of RAM and a Zotac Geforce GTX 1070 X 8G. Our \gls{ba} schema
is implemented on the GPU using CUDA 11. Since this work is focused on
global consistency, we perform our quantitative evaluation using the
RMSE on the absolute trajectory error (ATE) with $\bbSE(3)$
alignment. The metric's alignment is computed using the Horn
method~\cite{horn1988closed}, and the timestamps are used to determine
the associations. Then, we calculate the RMSE of the translational
differences between all matched poses.
In \secref{sec:rgbd}, we discuss the approaches and the datasets used for comparison with \rgbd~sensors,
while in \secref{sec:lidar} we present the results for \lidar~data.
Since our implementation can run both on GPU and CPU, we report the runtimes of our
algorithm for various pyramid resolutions using different commercial GPUs and our processor (\figref{fig:gpus}). The image reports timings for each complete optimization iteration at different resolutions. 

In our experiments, to switch from one level to the other, we use a simple termination criterion involving the variation of the error in \eqref{eq:total-error-multicue} over the iterations. The number of iterations for each level differs from the quality of the initial guess. In our experiments, we typically observe successful around 10 iterations on coarser, 5 on the middle, and just a couple on the finest level.



\subsubsection{RGB-D}
\label{sec:rgbd}
As a public benchmark for \rgbd~we used several sequences of ETH3D
\cite{schops2019bad}.  This dataset is acquired with global shutter
cameras and accurate active stereo depth. Modeling rolling shutter
effects and light changes cannot be encapsulated in the projection
function $\pi(\cdot)$, our only pipeline component that
differs between the \rgbd~and \lidar. For this reason, we restrict our
comparison to the setting mentioned above.

The work proposed in this paper refines the map from a reasonable
initial guess. We compute such initial configurations employing an
improved online CUDA version of MD-SLAM~\cite{di2022md}, our previous SLAM system that unifies depth sensors through hierarchical photometric odometry estimation and feature-based loop-closures.

We compare different approaches representative of different
classes of SLAM and \gls{ba} algorithms specific for \rgbd~sensors:
DVO-SLAM\cite{kerl2013dense},
ElasticFusion~\cite{whelan2015elasticfusion},
BundleFusion\cite{dai2017bundlefusion} BAD-SLAM\cite{schops2019bad},
ORB-SLAM2 \cite{mur2017orb}.
DVO SLAM implements a mixed geometry-based and direct
registration. Internally the alignment between pairs of keyframes is
obtained by jointly minimizing point-to-plane and photometric
residuals. This is similar to ElasticFusion, whose estimate consists
of a mesh model of the environment and the current sensor location
instead of the trajectory. BundleFusion refines the global estimate
by interleaving feature-based and photometric \gls{ba}. Similar to us, their
photometric refinement does not consider the structure, but
only the sensor poses. This method highly depends on data
association, employing correspondences based on sparse features and dense
geometric/photometric matching. BAD-SLAM is a surfel-based
direct Bundle Adjusted SLAM system that combines photometric and geometric errors
alternating optimization of motion and structure. In contrast to
these approaches, ORB-SLAM2 implements a traditional visual SLAM
pipeline, where a local map of landmarks around the RGB-D sensor is
constructed from ORB features \cite{rublee2011orb}. The map is
constantly optimized as the camera moves by performing local and global
\gls{ba}.

Most of the compared approaches run global refinement on a separate thread in an anytime fashion. The work presented in this paper
addresses only this global aspect. Reporting the timings of this experiment
would be unfair since our method addresses only a part of the problem.

ETH3D provides images of 740×460 pixels. We compute
a 3-level pyramid from these images with scales 1/2, 1/4, and 1/8. In
\tabref{tab:rgbd-eth}, we can see that our photometric refinement
performs on par (second after BAD-SLAM) with other state-of-the-art
ad-hoc RGB-D SLAM and \gls{ba} systems. Our method reduces the trajectory
error to a few millimeters. More importantly, it improves by $60\%$ the accuracy of
the initial guess provided by MD-SLAM. \figref{fig:hopt} and \figref{fig:augm-graph} illustrate the effect of the hierarchical optimization on self-recorded data.

Summarizing, using our general pipeline of MD-SLAM and photometric \gls{ba}
presented in this paper provides results comparable with other
\rgbd~specific approaches, being second only to BAD-SLAM.

\newcolumntype{L}{>{$}l<{$}}
\newcolumntype{C}{>{$}c<{$}}
\newcolumntype{R}{>{$}r<{$}}
\newcommand{\nm}[1]{\textnormal{#1}}

\begin{table} [t]
	\centering
	\tabcolsep=0.15cm
	\begin{tabular}{LCCCCCCC}
		\toprule
		&
		\multicolumn{1}{c}{\rotatebox[origin=c]{90}{ElasticFusion}} &
		\multicolumn{1}{c}{\rotatebox[origin=c]{90}{ORB-SLAM2}}&
		\multicolumn{1}{c}{\rotatebox[origin=c]{90}{DVO-SLAM}} &
		\multicolumn{1}{c}{\rotatebox[origin=c]{90}{BundleFusion}} &
		\multicolumn{1}{c}{\rotatebox[origin=c]{90}{BAD-SLAM}} &
		\multicolumn{1}{c}{\rotatebox[origin=c]{90}{MD-SLAM}} & 
		\multicolumn{1}{c}{\rotatebox[origin=c]{90}{MD-SLAM + \textbf{Ours}}} \\
		\midrule 
		
		\nm{table3}&-&0.007&0.008&0.017&\textbf{0.002}&0.016&0.009\\
		\nm{table4}&0.012&0.008&0.018&-&\textbf{0.002}&0.023&0.008\\
		\nm{table7}&-&0.010&0.007&0.010&\textbf{0.003}&0.018&0.009\\
		\nm{cables1}&0.018&0.007&\textbf{0.004}&0.022&0.007&0.021&0.006\\
		\nm{plant2}&0.017&0.003&0.003&0.004&\textbf{0.001}&0.005&\textbf{0.001}\\
		\nm{planar2}&0.011&0.005&\textbf{0.002}&0.003&0.003&0.009&0.004\\
		\midrule
		\nm{mean}&0.014&0.007&0.007&0.011&0.003&0.015&0.006\\
		\nm{std}&0.003&0.002&0.005&0.008&0.002&0.007&0.003\\
		\bottomrule
	\end{tabular}
	\caption{ATE RMSE [m] on ETH3D benchmark, recorded with global shutter camera and synchronous streams. ElasticFusion fails in \textit{table3} and \textit{table7}, BundleFusion fails in \textit{table4}.}
	\label{tab:rgbd-eth}
\end{table}

\subsubsection{3D LiDAR}
\label{sec:lidar}
To validate our approach on \lidar~measurements we used both our data and
public benchmarks.
We used the Newer College Dataset
\cite{zhang2021multicamera} as a public benchmark. The dataset is
recorded at 10 Hz with Ouster OS0-128. More specifically, we used the
\textit{cloister}, \textit{quad} (easy), and \textit{stairs} sequences.
The \emph{quad} sequence contains two loops that explore the
Oxford campus courtyard, \emph{cloister} mixes outdoor and indoor
scenes while \emph{stairs} captures an indoor scenario with multiple
floors. Being based on image comparison, our approach operates well
on \lidar~data having a good vertical resolution. \lidars~ with fewer
beams (i.e., 64, 32, 16) would produce an unbalanced horizontal image,
reducing the converge basin of the algorithm.

We compare our method with BALM2 \cite{liu2021balm}, which, to the best of our knowledge, is the only publicly available \lidar~global refinement approach.
BALM2 is based on the overall consistency of points, lines, and planes.
This system requires the same input as our method, namely
an initial guess trajectory and the point clouds.

To compute the initial guess, we used several SLAM algorithms specific for
\lidar: LeGO-LOAM~\cite{shan2018lego}, SuMA~\cite{behley2018efficient} and our
unified MD-SLAM.
LeGO-LOAM is a
pure geometric feature-based frame-to-model LiDAR SLAM system, where
the optimization on roll, yaw, and z-axis (pointing up) is decoupled
from the planar parameters. SuMa constructs a surfel-based map and
estimates the changes in the sensor's pose by exploiting the
projective data association in a frame-to-model or frame-to-frame
fashion.

\tabref{tab:lidar-ncc} reports the accuracy of our method and BALM2,
for each dataset, and each initial guess. The ATE of the SLAM solution measures the quality of an initial guess. We observe that
LeGO-LOAM provides a good guess on all planar data but fails on the
\emph{stairs} dataset resulting in an ATE of more than 3 meters. MD-SLAM performs reasonably well with a maximum ATE of 0.36 meters, while SuMA
yields an acceptable initial guess only in the stairs dataset. If the initial guess is good, both BALM2 and our method perform well,
improving the initial estimate. However, as the initial guess
degrades, our unified global refinement's accuracy remains stable
by systematically improving the trajectory estimate. On these data, we
observed BALM2 to be particularly sensible to rotations and less dense
trajectories (i.e., trajectories sampled with fewer keyframe poses). Quantitative results show that our strategy is
successful within \lidar~data, raising the initial estimate close to
60\% improvement when a good initial guess is provided (i.e., \textit{stairs} with MD-SLAM).

\begin{table} [h]
	\centering
	\tabcolsep=0.15cm
	\begin{tabular}{LCCC|CCC|CCC}
		\toprule
		\multicolumn{1}{l}{} &
		\multicolumn{3}{c}{LeGO-LOAM} &
		\multicolumn{3}{c}{MD-SLAM} &
		\multicolumn{3}{c}{SuMA} \\
		\cmidrule{2-10}
		&
		\multicolumn{1}{c}{\rotatebox[origin=c]{90}{SLAM}} &
		\multicolumn{1}{c}{\rotatebox[origin=c]{90}{BALM2}}&
		\multicolumn{1}{c|}{\rotatebox[origin=c]{90}{\textbf{Ours}}} &
		\multicolumn{1}{c}{\rotatebox[origin=c]{90}{SLAM}} & 
		\multicolumn{1}{c}{\rotatebox[origin=c]{90}{BALM2}} &
		\multicolumn{1}{c|}{\rotatebox[origin=c]{90}{\textbf{Ours}}} &
		\multicolumn{1}{c}{\rotatebox[origin=c]{90}{SLAM}} &
		\multicolumn{1}{c}{\rotatebox[origin=c]{90}{BALM2}} &
		\multicolumn{1}{c}{\rotatebox[origin=c]{90}{\textbf{Ours}}}\\
		\midrule 
		\nm{cloister}&0.20&0.25&\textbf{0.16}&0.36&0.37&\textbf{0.32}&3.34&2.67&\textbf{2.53}\\
		\nm{quad}&\textbf{0.09}&\textbf{0.09}&\textbf{0.09}&0.25&1.98&\textbf{0.17}&1.74&1.72&\textbf{1.68}\\
		\nm{stairs}&\textbf{3.20}&5.13&3.48&0.23&0.38&\textbf{0.10}&0.67&0.86&\textbf{0.60}\\
		\bottomrule
	\end{tabular}
	\caption{ATE RMSE [m] on Newer College Dataset, recorded with OS0-128. We always improve the SLAM baseline, a part for \textit{stairs} starting from LeGO-LOAM estimate.}
	\label{tab:lidar-ncc}
\end{table}

The convergence basin of these global refinement strategies
are bounded by inconsistency in the laser measurements (i.e., skewed point clouds); this is why none of the systems can further enhance the trajectory in the LeGO-LOAM \textit{quad} experiment.

\figref{fig:motiv} shows our large-scale reconstruction of the
historical part of Viterbo (Italy) from self-recorded data using a hand-held Ouster OS0-128 \lidar. The trajectory is about 2 km long, and the dataset is available from our repository \footref{repo}.

\subsubsection{RGB-D + 3D LiDAR}
\label{sec:fusion}
Thanks to the unified nature of our pipeline, it allows 
the straightforward integration of multiple heterogeneous sensors. In
\secref{sec:error-min}, we formulated an optimization problem that
estimates the sensor's trajectory $\bX_{1:N}$.

In a multiple sensors configuration, we can express the optimization
problem as a function of the trajectory of a multi-device platform, to
which all sensors are rigidly attached. The multiple sensors setting slightly modify the cost function presented in ~\eqref{eq:total-error-multicue} to
include the constant sensor offsets. Our supplementary material \footref{repo} reports this extended formulation. The cost function for multi-sensor
photometric refinement is just the sum of the cost functions of each device.
In this section, we report an analysis of our optimization strategy with both RGB-D and LiDAR. 
For these experiments, we mounted an Intel D455 on
the top of an OS0-128 and accurately calibrate the offset between the two sensors using a robust point-to-plane alignment \cite{chen1992object}. We
recorded some static indoor sequences, each of them containing
synchronized grayscale and depth images from D455 and its
corresponding intensity and range images from the OS0-128.

We carried out an experiment to evaluate the accuracy and converge basin of our approach in this configuration. To this extent, we perform our
photometric alignment strategy using a single sensor frame consisting
of a \rgbd~ and a \lidar~ measurement pair. Since we align a
sensor frame with respect to itself, we expect the computed estimate to be as
close as possible to the identity. The optimization is carried on
starting from increasingly wrong initial guesses. 

In \figref{fig:fusion}, we show the result of this experiment, starting from single sensor optimization as a baseline. Due to their different characteristics, the two sensors behave very differently
during photometric alignment. The 3D \lidar~has an inferior
resolution but a very large $360 \deg$ horizontal FoV. This
results in a relatively large convergence basin, but the limited
resolution affects the minimum. Conversely, \rgbd, provides a high-resolution image with a much smaller FoV of around $70 \deg$. This
results in better minima at the expense of a smaller converge basin.

We can refine the data in two ways that we name \textit{coupled} and
\textit{consecutive}.  The coupled approach aims to simultaneously
finding the minimum of both photometric errors of
\rgbd~$F^\mathrm{r}(\XSet)$ and \lidar~$F^\mathrm{l}(\XSet)$, as follows
\begin{equation}
\XSet^\mathrm{coupled} = \argmin_{\XSet} F^\mathrm{r}(\XSet) + F^\mathrm{l}(\XSet)
\end{equation}
The \textit{consecutive} optimization combines the strengths of the two
sensors, and operates by first carrying on an optimization where only the \lidar~is used.
Subsequently, it uses the solution of this first run to find a minimum for the \rgbd~problem.

The results suggest that fusing the two sensors always performs better
than single sensors (\figref{fig:fusion}). Specifically,
\textit{coupled} is generally more accurate compared to \rgbd~only,
having the converge basin incremented by the \lidar. Similarly,
\textit{consecutive} seems to be the best since it takes full advantage of the large convergence basin of the \lidar~and benefits from the resolution of \rgbd.
We believe that this proof of concept opens ways for 3D
reconstruction algorithms that symmetrically fuse the two sensors, taking benefits from both.


\balance

\section{Conclusion}
\label{sec:conclusion}

In this paper, we presented a photometric \gls{ba} methodology that operates both with \rgbd~and \lidar~in a unified manner. To the best of our knowledge, our approach is the only open-source \gls{ba} system that works independently for depth sensors and specifically exploits the photometric capabilities of the \lidar~image. Comparative experiments show that our simple schema performs on par or better compared to existing sensor-specific state-of-the-art approaches without making any assumption about the environment and free from data association. Thanks to our photometric registration methodology's inherent data separation, we develop our software in CUDA and release an open-source implementation. Considering the larger convergence basin and accuracy obtained fusing both \rgbd~and \lidar~within our registration strategy, we envision a SLAM/\gls{ba} pipeline that uses both depth sensors jointly. Furthermore, to complete our universal 3D reconstruction schema, future research would involve finding generic structure representation to optimize both \rgbd~and \lidar~measurements.

\bibliographystyle{plain}
\bibliography{robots}


\onecolumn

\section*{Supplementary Material}
In this section, we explain in detail our photometric error term presented in \secref{eq:total-error-multicue} and provide analytic expressions for the Jacobian matrices. In addition, we illustrate some additional trajectory plots related to the \lidar{} experiments. We did not provide the same for \rgbd~since trajectory errors are too low. 

We represent as $\bDelta\bx$ the Lie algebra
$\lieSE(3)$ associated with the group $\bbSE(3)$, parameterized as $\bDeltax = [\bDelta\bt, \bDelta\bq]^T$.
$\bDelta\bt \in \bbR^3$ is the translation, and $\bDelta\bq \in \bbR^3$ is the imaginary part of a unit quaternion.  The rotation matrix can be calculated from the perturbation vector using the \textit{exponential map} at the identity $\bDeltaR = \exp(\bDelta\bq)$. We extend the notation of the exponential map to refer to the transformation encoded in $\bDeltax$. Let $\bDeltaX=\exp(\bDeltax)$, be this transformation whose rotation is $\bDeltaR$ and translation $\bDeltat$. We use the $\boxplus$ operator to denote applying a perturbation to a transform  $\bX \exp (\bDeltax) := \bX \boxplus \bDeltax$.

In the reminder, the error term differs slightly from the one presented in \eqref{eq:total-error-multicue}. Here we insert the offset mapping \rgbd{} or \lidar{} with respect to the reference frame of a multi-device sensor platform. This is fundamental to fuse the two sensors as illustrated in \secref{sec:fusion}. The constant rigid transformation comprises rotation $\bR_o$ and translation $\bt_o$.  

For compactness, we define two quantities $\bp_u$ as the point transformed by this offset and $\bar \bp_u$ as the point $\bp_u$ transformed by the estimate and the inverse of the offset as follows:

\newcommand{\bpu}{\bR_o \pi^{-1} \left(\bu,\depth \right) + \bt_o}

\newcommand{\icp}{ \bR_o^T \left( \bR_j^T \left ( \bR_i \bp_u + \bt_i   - \bt_j \right ) -\bt_o \right ) }

\newcommand{\perti}{ \bR_o^T \left( \bR_j^T \left ( \bR_i \left ( \bR(\bDelta\bq) \bp_u + \bDelta\bt \right ) + \bt_i   - \bt_j \right ) -\bt_o \right ) }

\newcommand{\pertj}{\bR_o^T \left (  \bR(-\bDelta\bq) \bR_j^T \left ( \bR_i  \bp_u  + \bt_i   - \bt_j \right ) - \bDelta \bt -\bt_o \right ) }

\begin{align}
\bp_u &= \bpu, \\
\bar \bp_u &= \icp .
\end{align}
Thus, we can rewrite our error in the following way:
\begin{align}
\be_u^c &= \map^\channel( \bX_{j,i},\I_i^\channel(\bu)) - \I_j^\channel(\bu') = \map^\channel \left ( \bar \bp_u \right ) - \I_j^\channel \left ( \pi \left ( \bar \bp_u \right ) \right).
\label{eq:error-jacobian}
\end{align}
Applying the perturbations $\bDelta\bx_i$ and $\bDelta\bx_j$ on the right hand side in \eqref{eq:error-jacobian} leads to:
\begin{align}
\be_u^c (\bX_i \boxplus \bDeltax_i) &=\map^\channel \left ( \bar \bp_{u|i} \right ) - \I_j^\channel \left ( \pi \left ( \bar \bp_{u|i} \right ) \right ),\\
\be_u^c (\bX_j \boxplus \bDeltax_j) &=\map^\channel \left(  \bar \bp_{u|j} \right) - \I_j^\channel \left ( \pi \left ( \bar \bp_{u|j} \right ) \right ).
\end{align}
with $\bar \bp_{u|i}$ and $\bar \bp_{u|j}$ respectively defined as:
\begin{align}
	\bar \bp_{u|i} &= \perti,\\
	\bar \bp_{u|j} &= \pertj .
\end{align}
Deriving these two quantities with respect to the perturbations leads to the following Jacobians:
\begin{align}
\frac{\partial \bar \bp_{u|i}}{\partial \bDeltax_i} &= \bR_o^T \bR_j^T \bR_i
\begin{bmatrix}
\bI_{3 \times 3} & 2 \skew{\bp_u}
\end{bmatrix},\\
\frac{\partial \bar \bp_{u|j}}{\partial \bDeltax_j} &= -\bR_o^T
\begin{bmatrix}
\bI_{3 \times 3} & 2 \skew{ \bR_j^T \left ( \bR_i \bp_u + \bt_i - \bt_j \right ) }
\end{bmatrix}.
\end{align}
Projective Jacobians depends on the projection model, differing \rgbd{} and \lidar{}, these are calculated respectively deriving \eqref{eq:pinhole-projection} and \eqref{eq:spherical-projection} with respect to the transformed point $\bar \bp_u$.\\

\noindent\textbf{Pinhole model:}
\begin{equation}
\frac{\partial \pi_p (\bar\bp_u)}{\partial \bar\bp_u} = \frac{\partial \phi (\bK \bar\bp_u)}{\partial \bar\bp_u} = 
\frac{1}{v_z^2} 
\begin{bmatrix}
v_z & 0 & -v_x\\
0 & v_z & -v_y
\end{bmatrix} \Big |_{\left [ v_x, v_y, v_z \right ] = \bK \bar\bp_u} \bK.
\end{equation}

\noindent\textbf{Spherical model:}
\begin{equation}
\frac{\partial \pi_s (\bar\bp_u)}{\partial \bar\bp_u} = \frac{\partial \bK_{[1,2]}  \psi(\bar\bp_u)}{\partial \bar\bp_u} = \bK_{[1,2]}
\begin{bmatrix}
\frac{1}{v_x^2 + v_y^2}\begin{bmatrix}-v_y & v_x & 0\end{bmatrix}\\
\frac{1}{v_x^2 + v_y^2 + v_z^2}\,\begin{bmatrix}-\frac{v_x v_z}{\sqrt{v_x^2 + v_y^2}} & -\frac{v_y v_z}{\sqrt{v_x^2 + v_y^2}} & \sqrt{v_x^2 + v_y^2}\end{bmatrix}\\
\begin{bmatrix}0 & 0 & 0\end{bmatrix}
\end{bmatrix} \Big |_{\left [ v_x, v_y, v_z \right ] = \bar\bp_u}.
\end{equation}
Image jacobians are numerically computed for each channel $c$ with pixel-wise derivation:
\begin{align}
\frac{\partial \I^c_{r,c}}{\partial r} &= \frac{1}{2}\left(\I^c_{r+1,c} - \I^c_{r-1,c}\right) \nonumber \\
\frac{\partial \I^c_{r,c}}{\partial c} &= \frac{1}{2}\left(\I^c_{r,c+1} - \I^c_{r,c-1}\right)
\end{align}
\\~\\
Jacobians on the mapping function $\map^\channel$ for each channel grayscale/intensity, range/depth and normals, respectively $\{ \mathrm{g},\mathrm{\depth},\mathrm{n} \}$ are computed also with respect to perturbations, since the estimates appears also to the left of the error function \eqref{eq:error-jacobian}. In the reminder we specifically write mapping function and Jacobians of each cue. 

\vspace{1cm}

\noindent\textbf{Intensity}\\
The mapping function does not affect the intensity/grayscale channel, therefore we have:
\begin{align}
\map^\mathrm{g}(\bar \bp_{u}) = \I^\mathrm{g}(\bar \bp_{u}), \quad \quad \frac{\partial \map^\mathrm{g}(\bar \bp_{u|i})}{\bDelta\bx_i} = 0, \quad \quad \frac{\partial \map^\mathrm{g}(\bar \bp_{u|j})}{\bDelta\bx_j} = 0.
\end{align}

\vspace{1cm}
\noindent We need to differ between the two depth sensor models, since \rgbd{} provides depth while \lidar{} range measurements. Hence, leading to different Jacobians.\\

\noindent\textbf{Depth}

\begin{align}
\map^{\mathrm{d}}(\bar \bp_{u}) = \begin{bmatrix} 0 & 0 & 1 \end{bmatrix} \bar \bp_{u}, \quad \quad \frac{\partial \map^{\mathrm{d}}(\bar \bp_{u|i})}{\bDelta\bx_i}  = \begin{bmatrix} 0  & 0 & 1 \end{bmatrix} \frac{\partial \bar \bp_{u|i}}{\partial \bDelta \bx_i}, \quad \quad \frac{\partial \map^{\mathrm{d}}(\bar \bp_{u|j})}{\bDelta\bx_j}  = \begin{bmatrix} 0 & 0 & 1 \end{bmatrix} \frac{\partial \bar \bp_{u|j}}{\partial \bDelta\bx_j}.
\end{align}

\noindent\textbf{Range}
\begin{align}
\map^{\mathrm{\depth}}(\bar \bp_{u}) = || \bar \bp_{u} ||,  \quad \quad  \frac{\partial \map^{\mathrm{\depth}}(\bar \bp_{u|i})}{\bDelta\bx_i} =\frac{\bar\bp_u^T}{||\bar\bp_u||} \frac{\partial \bar \bp_{u|i}}{\partial \bDelta\bx_i}, \quad \quad
\frac{\partial \map^{\mathrm{\depth}}(\bar \bp_{u|j})}{\bDelta\bx_j} = \frac{\bar\bp_u^T}{||\bar\bp_u||} \frac{\partial \bar \bp_{u|j}}{\partial \bDelta\bx_j}.\\
\end{align}

\vspace{1cm}

\noindent\textbf{Normals}\\
Normals, differently, affect only rotation, hence is convenient to rewrite our error function expressed in \eqref{eq:error-jacobian}. This reduces to:
\begin{align}
\be_u^{\mathrm{n}} &= \map^\mathrm{n} \left ( \bR_o^T \bR_j^T \bR_i \bR_o \bn_u \right ) - \I^\mathrm{n} \left ( \pi \left ( \bar \bp_u \right ) \right).
\label{eq:normal-error-jacobian}
\end{align}
It is trivial to derive \eqref{eq:error-jacobian} with respect to the angular parts of the perturbations. Note that $\bn_u$ is the normal prior to any transformation, since everything is enrolled in \eqref{eq:normal-error-jacobian}.

\begin{align}
\frac{\partial \map^{\mathrm{n}} \left ( \bR_o^T \bR_j^T \bR_i \bR(\bDelta \bq_i) \bR_o \bn_u \right) }{\bDelta\bq_i} &= 2\bR_o^T \bR_j \bR_i \bR_o \skew{\bn_u},\\ 
\frac{\partial \map^{\mathrm{n}} \left ( \bR_o^T \bR(- \Delta\bq_j) \bR_j^T \bR_i \bR_o \bn_u \right )}{\bDelta\bq_j} 
&= -2\bR_o^T \skew{\bR^T_j \bR_i \bR_o \bn_u}.
\end{align}

\vspace{1cm}

\noindent Finally, we can reconstruct the full jacobians:
\begin{align}
	\bJ_i &= \frac{\partial \map^{\mathrm{c}}(\bar \bp_{u|i})}{\bDelta\bx_i} - \frac{\partial \I^c_{r,c}}{\partial r, c} \frac{\partial \pi (\bar\bp_u)}{\partial \bar\bp_u} \frac{\partial \bar \bp_{u|i}}{\partial \bDeltax_i}, \\
	\bJ_j &=\frac{\partial \map^{\mathrm{c}}(\bar \bp_{u|j})}{\bDelta\bx_j} - \frac{\partial \I^c_{r,c}}{\partial r, c} \frac{\partial \pi (\bar\bp_u)}{\partial \bar\bp_u} \frac{\partial \bar \bp_{u|j}}{\partial \bDeltax_j}. 	
\end{align}

\begin{figure}[t]
	\begin{subfigure}{0.99\linewidth}
		\centering
		\includegraphics[width=\textwidth]{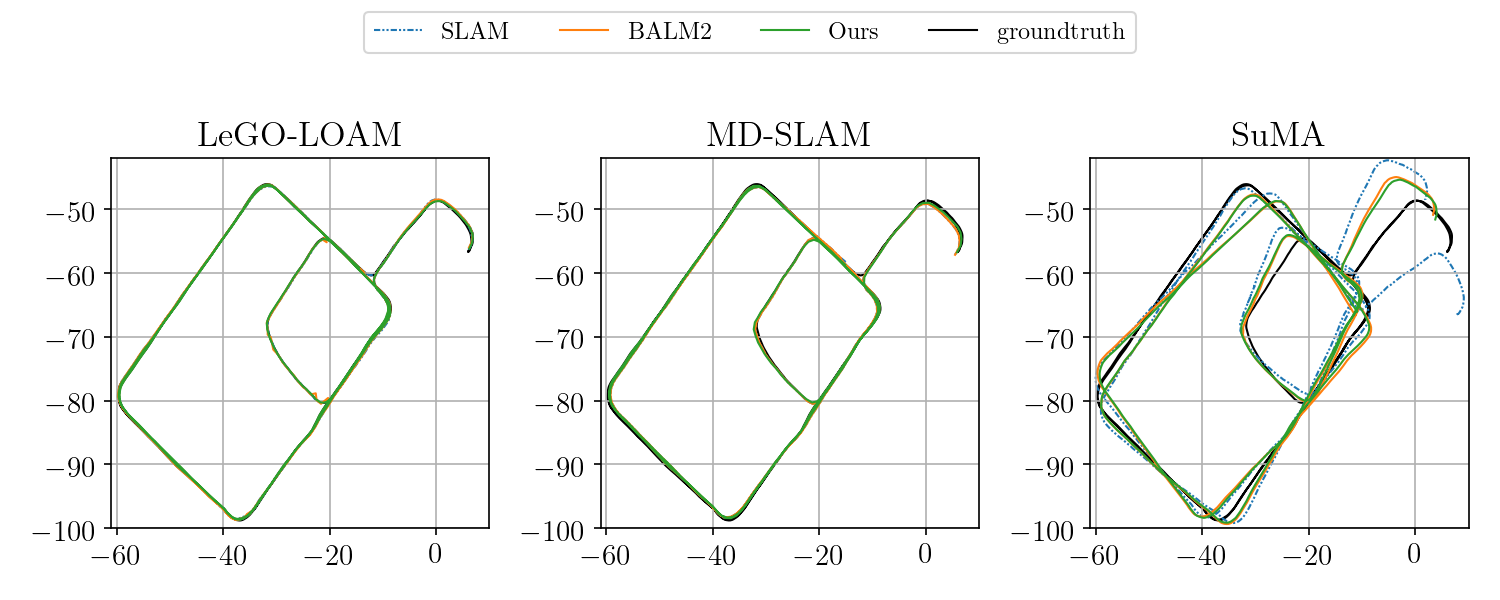}
		\caption{Trajectory estimates of \textit{cloister} sequence of Newer College Dataset. SLAM estimate in blue, BALM2 estimate in orange, Ours in green and groundtruth in black.}
		\label{fig:cloister}
	\end{subfigure}
	\hfill
	\begin{subfigure}{0.99\linewidth}
		\centering
		\includegraphics[width=\textwidth]{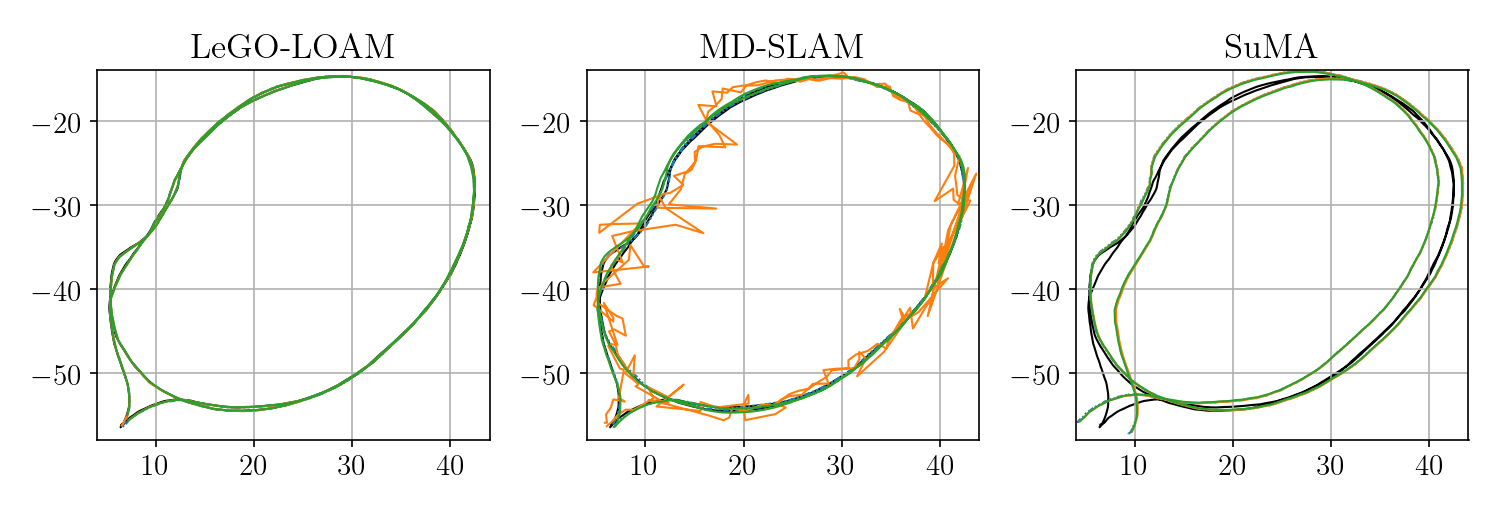}
		\caption{Trajectory estimates of \textit{quad} sequence of Newer College Dataset. SLAM estimate in blue, BALM2 estimate in orange, Ours in green and groundtruth in black. BALM2 suffer of less dense trajectories, since MD-SLAM spawn less keyframe poses  to reduce odometry drift.}
		\label{fig:quad}
	\end{subfigure}
	\begin{subfigure}{0.99\linewidth}
		\hspace{0.15cm}
		\includegraphics[width=\textwidth]{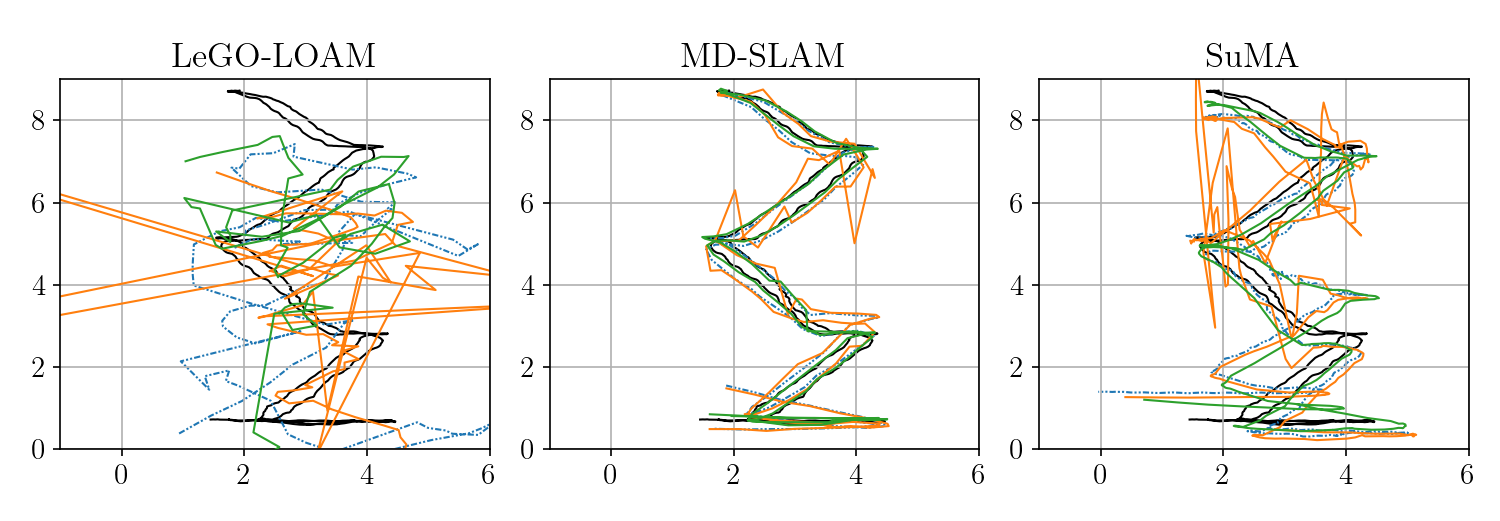}
		\caption{Trajectory estimates of \textit{stairs} sequence of Newer College Dataset. SLAM estimate in blue, BALM2 estimate in orange, Ours in green and groundtruth in black. BALM2 suffer of rotations, as it is observable from MD-SLAM and SuMA initial guess. For completeness we include the LeGO-LOAM plot, however both our and BALM2 fail due to bad initial guess.}
		\label{fig:stairs}
	\end{subfigure}
\end{figure} 

\twocolumn

\balance

\end{document}